\documentclass[conference]{IEEEtran}
\IEEEoverridecommandlockouts

\usepackage{amsmath}
\usepackage{amssymb}    
\usepackage{times,amsmath,amssymb,amsfonts,amsthm}
\usepackage[ruled,vlined]{algorithm2e}
\usepackage{graphicx}
\usepackage{tikz}
\usetikzlibrary{graphs}
\usepackage{mathtools}
\usepackage[all]{xy}
\usepackage{psfrag}

\newtheorem*{theorem*}{Theorem}

\theoremstyle{definition}
\newtheorem{example}{Example}

\newcommand{\R}{\mathbb{R}}

\newcommand{\vetx}{\boldsymbol{x}}
\newcommand{\vety}{\boldsymbol{y}}

\newcommand{\bb}{\begin{equation}}
\newcommand{\ee}{\end{equation}}
\newcommand{\bbb}{\begin{eqnarray}}
\newcommand{\eee}{\end{eqnarray}}
\newcommand{\benu}{\begin{enumerate}}
\newcommand{\eenu}{\end{enumerate}}

\newcommand{\vetu}{\boldsymbol{u}}
\newcommand{\vetw}{\boldsymbol{w}}
\newcommand{\vetv}{\boldsymbol{v}}

\newcommand{\bpm}{\begin{bmatrix}}
\newcommand{\epm}{\end{bmatrix}}

\newcommand{\sprod}[2]{\left\langle #1, #2 \right\rangle}

\newcommand{\ii}{\mathbf{i}}
\newcommand{\jj}{\mathbf{j}}
\newcommand{\kk}{\mathbf{k}}
\newcommand{\quat}[1]{{#1}_0 + {#1}_1 \ii + {#1}_2 \jj + {#1}_3 \kk}
\newcommand{\re}[1]{\text{Re}\left\{#1\right\}}
\newcommand{\ve}[1]{\text{Ve}\left\{#1\right\}}

\newcommand{\qeq}{\quad \mbox{and} \quad}

\def\boxmax{\kern 0em\hbox{\rm \kern .25em\lower.1ex\hbox{\rlap{$\vee$}}\kern -.07em\lower.2ex\hbox{$\square$}\kern.25em}}
\def\boxmin{\kern 0em\hbox{\rm \kern .25em\lower.1ex\hbox{\rlap{$\wedge$}}\kern -.07em\lower.2ex\hbox{$\square$}\kern.25em}}
\def\boxdiamond{\kern 0em\hbox{\rm \kern .25em\hbox{\rlap{$\diamond$}}\kern -.15em\lower.2ex\hbox{$\square$}}}

\def\BibTeX{{\rm B\kern-.05em{\sc i\kern-.025em b}\kern-.08em
    T\kern-.1667em\lower.7ex\hbox{E}\kern-.125emX}}
\begin{document}

\newpage
\thispagestyle{empty}
\begin{minipage}{0.8\textwidth}
\noindent {\Huge  IEEE Copyright Notice} \vspace{2cm}

\noindent $\textcopyright$ 2019 IEEE.  Personal use of this material is permitted.  Permission from IEEE must be obtained for all other uses, in any current or future media, including reprinting/republishing this material for advertising or promotional purposes, creating new collective works, for resale or redistribution to servers or lists, or reuse of any copyrighted component of this work in other works.
\vspace{1cm}

\noindent Accepted to be Published in: Proceedings of the 8th Brazilian Conference on Intelligent Systems (BRACIS 2019), October 15-18, 2019, Salvador, BA, Brazil.
\end{minipage}

\title{An Introduction to Quaternion-Valued Recurrent Projection Neural Networks \\
\thanks{This work was supported in part by CNPq under grant no. 310118/2017-4, FAPESP under grant no. 2019/02278-2, and Coordena\c{c}\~ao  de Aperfei\c{c}oamento  de Pessoal de N\'ivel Superior - Brasil (CAPES) - Finance Code 001.}
}

\author{
\IEEEauthorblockN{Marcos Eduardo Valle and Rodolfo Anibal Lobo}
\IEEEauthorblockA{\textit{Institute of Mathematics, Statistics, and Scientific Computing} \\
\textit{University of Campinas}\\
Campinas, Brazil \\
valle@ime.unicamp.br, rodolfolobo@ug.uchile.cl
}
}

\maketitle

\begin{abstract}
Hypercomplex-valued neural networks, including quaternion-valued neural networks, can treat multi-dimensional data as a single entity. In this paper, we introduce the quaternion-valued recurrent projection neural networks (QRPNNs). Briefly, QRPNNs are obtained by combining the non-local projection learning with the quaternion-valued recurrent correlation neural network (QRCNNs). We show that QRPNNs overcome the cross-talk problem of QRCNNs. Thus, they are appropriate to implement associative memories. Furthermore, computational experiments reveal that QRPNNs exhibit greater storage capacity and noise tolerance than their corresponding QRCNNs. 
\end{abstract}

\begin{IEEEkeywords}
Recurrent neural network, Hopfield network, associative memory, quaternion-valued neural network.
\end{IEEEkeywords}

\section{Introduction}
The Hopfield neural network, developed in the early 1980s, is an important and widely-known recurrent neural network which can be used to implement associative memories \cite{hopfield82,hopfield85}. Successful applications of the Hopfield network include control \cite{gan17,song17}, computer vision and image processing \cite{wang15,jli16}, classification \cite{pajares10,zhang17}, and optimization \cite{hopfield85,serpen08,cli16}. 

Despite its many successful applications, the Hopfield network may suffer from a very low storage capacity when used to implement associative memories. Precisely, due to cross-talk between the stored items, the Hebbian learning adopted by Hopfield in his original work allows for the storage of approximately $n/(2\ln n)$ items, where $n$ denotes the length of the stored vectors \cite{mceliece87}. 

Several neural networks and learning rules have been proposed in the literature to increase the storage capacity of the original bipolar Hopfield network. For example, Personnaz et al. \cite{personnaz85} as well as Kanter and Sompolinsky \cite{kanter87} proposed the projection rule to determine the synaptic weights of the Hopfield networks. The projection rule increases the storage capacity of the Hopfield network to $n-1$ items. Another simple but effective improvement on the storage capacity of the original Hopfield networks was achieved by Chiueh and Goodman's recurrent correlation neural networks (RCNNs) \cite{chiueh91,chiueh93}. Briefly, an RCNN is obtained by decomposing the Hopfield network with Hebbian learning into a two layer recurrent neural network. The first layer computes the inner product (correlation) between the input and the memorized items followed by the evaluation of a non-decreasing continuous activation function. The subsequent layer yields a weighted average of the stored items. Alternatively, certain RCNNs can be viewed as kernelized versions of the Hopfield network with Hebbian learning \cite{moreno04,perfetti08}. 

It turns out that the associative memory models described in the previous paragraphs are all designed for the storage and recall of bipolar real-valued vectors. In many applications, however, we have to process multivalued or multidimensional data \cite{hirose12}. In view of this remark, the Hopfield neural network as well as the RCNNs have been extended to hypercomplex systems such as complex numbers and quaternions. 

Research on complex-valued Hopfield neural networks dates to the late 1980s \cite{noest88a,noest88b,aizenberg92}. In 1996, Jankowski et al. \cite{jankowski96} proposed a multistate complex-valued Hopfield network with Hebbian learning that corroborated to the development of many other hypercomplex-valued networks. Namely, Lee developed the projection rule for (multistate) complex-valued Hopfield networks \cite{lee06}. Based on the works of Jankowski et al. and Lee, Isokawa et al. proposed a multistate quaternion-valued Hopfield neural network using either Hebbian learning or projection rule \cite{isokawa13}. In 2014, Valle proposed a complex-valued version of the RCNNs \cite{valle14nnB}. Recently, the RCNNs have been further extended to quaternions \cite{valle18wcci}. 

At this point, we would like call the reader's attention to the following: Preliminary experiments revealed that hypercomplex-valued recurrent neural networks, interpreted as a dynamical systems, are less susceptible to chaotic behavior than their corresponding real-valued network \cite{castro19arxiv}. For example, hypercomplex-valued Hopfield networks usually require fewer updates to settle down at an equilibrium state than their corresponding real-valued neural networks. In view of this remark, we shall focus on quaternion-valued recurrent neural networks that can be used to implement associative memories. 

Precisely, in this paper we propose an improved version of the quaternion-valued RCNNs \cite{valle18wcci}. Although real, complex, and quaternion-valued RCNNs can be used to implement high-capacity associative memories, they require a sufficiently large parameter which can be impossible in practical implementations \cite{chiueh93,valle18wcci}. In order to circumvent this problem, in this paper we combine the projection rule and the RCNNs to obtain the new recurrent projection neural networks (RPNNs). As we will show, RPNNs always have optimal absolute storage capacity. Furthermore, the noise tolerance of RPNNs are usually higher than their corresponding RCNNs.  

The paper is organized as follows: Next section presents some basic concepts on quaternions. A brief review on the quaternion-valued Hopfield neural network (QHNN) and quaternion-valued recurrent correlation neural networks (QRCNNs) are given respectively in Sections \ref{sec:QHNN} and \ref{sec:QRCNNs}. Quaternion-valued recurrent projection neural networks (QRPNNs) are introduced in Section \ref{sec:QRPNNs}. The paper finishes with the concluding remarks in Section \ref{sec:concluding}.

%This document is a model and instructions for \LaTeX.
%Please observe the conference page limits. 

\section{Some Basic Concepts on Quaternions} \label{sec:quaternions}

Quaternions are hyper-complex numbers that extend the real and complex numbers systems.
A quaternion may be regarded as a 4-tuple of real numbers, i.e., $q=(q_0,q_1,q_2,q_3)$. Alternatively, a quaternion $q$ can be written as follows
\bb q = q_0 + q_1 \ii + q_2 \jj + q_3 \kk, \label{eq:quaterion} \ee
where $\ii, \jj$, and $\kk$ are imaginary numbers that satisfy the following identities:
\bb \ii^2 = \jj^2 = \kk^2 = \ii \jj \kk = -1. \ee
Note that $1,\ii,\jj$, and $\kk$ form a basis for the set of all quaternions, denoted by $\mathbb{H}$. 

A quaternion $q = \quat{q}$ can also be written as $q = q_0+\vec{q}$, where $q_0$ and $\vec{q}=q_1 \ii + q_2 \jj + q_3 \kk$ are called respectively the real part and the vector part of $q$. The real and the vector part of a quaternion $q$ are also denoted by $\re{q}:=q_0$ and $\ve{q}:=\vec{q}$.

The sum $p+q$ of two quaternions $p = \quat{p}$ and $q=\quat{q}$ is the quaternion obtained by adding their components, that is,
\bb \label{eq:sum} p+q = (p_0+q_0)+(p_1+q_1) \ii + (p_2+q_2) \jj + (p_3+q_3) \kk. \ee
Furthermore, the product $pq$ of two quaternions $p = p_0 + \vec{p}$ and $q = q_0 + \vec{q}$ is the quaternion given by 
\bb \label{eq:product} pq = p_0 q_0 - \vec{p} \cdot \vec{q} +p_0 \vec{q} + q_0 \vec{p} + \vec{p} \times \vec{q},\ee
where $\vec{p} \cdot \vec{q}$ and $\vec{p} \times \vec{q}$ denote respectively the scalar and cross products commonly defined in vector algebra. Quaternion algebra are implemented in many programming languages, including {\tt MATLAB}, {\tt GNU Octave}, {\tt Julia}, and {\tt python}.
We would like to recall that the product of quaternions is not commutative. Thus, special attention should be given to the order of the terms in the quaternion product. 

The conjugate and the norm of a quaternion $q=q_0+\vec{q}$, denoted respectively by $\bar{q}$ and $|q|$, are defined by 
\bb \label{eq:conjugate} \bar{q} = q_0-\vec{q} \qeq 
|q| = \sqrt{\bar{q} q} =  \sqrt{q_0^2 + q_1^2 + q_2^2 +q_3^2}. \ee
%Recall that $\overline{(\bar{q})}=q$, $\overline{(p+q)} = \bar{p} + \bar{q}$, and $\overline{(pq)} = \bar{q} \bar{p}$ hold true for any $p,q \in \mathbb{H}$.
We say that $q$ is a {\em unit} quaternion if $|q|=1$. We denote by $\mathbb{S}$ the set of all unit quaternions, i.e., $\mathbb{S} = \{q \in \mathbb{H}: |q|=1|\}$. Note the $\mathbb{S}$ can be regarded as an hypersphere in $\mathbb{R}^4$. The quanternion-valued function $\sigma:\mathbb{H}^* \to \mathbb{S}$ given by 
\bb \label{eq:sigma} \sigma(q) = \frac{q}{|q|}, \ee
maps the set of non-zero quaternions $\mathbb{H}^* = \mathbb{H} \setminus\{0\}$ to the set of all unit quaternions. The function $\sigma$ can be interpreted as a generalization of the signal function to unit quaternions.

Finally, the inner product of two quaternion-valued column vectors $\vetx=[x_1,\ldots,x_n]^T \in \mathbb{H}^n$ and $\vety = [y_1,\ldots,y_n]^T \in \mathbb{H}^n$ is given by 
\bb \label{eq:inner} \sprod{\vetx}{\vety} = \sum_{i=1}^n \bar{y}_i x_i. \ee
In particular, we have
\bb \re{\sprod{\vetx}{\vety}} = \sum_{i=1}^n (y_{i0}x_{i0} + y_{i1}x_{i1} + y_{i2}x_{i2} + y_{i3}x_{i3}),\ee
which corresponds to the inner product of the real-valued vectors of length $4n$ obtained by concatenating the 4-tuples $(x_{i0},x_{i1},x_{i2},x_{i3})$ and $(y_{i0},y_{i1},y_{i2},y_{i3})$, for $i=1,\ldots,n$. 

\section{Quaternion-Valued Hopfield Neural Networks} \label{sec:QHNN}

The famous {\em Hopfield neural network} (HNN) is a recurrent model which can be used to implement associative memories \cite{hopfield82}. Quaternion-valued versions of the Hopfield network, which generalize complex-valued models, have been extensively investigated in the past years \cite{isokawa06,isokawa07,isokawa08,isokawa12,osana12}. A comprehensive review on several types of {\em quaternionic HNN} (QHNN) can be found in \cite{isokawa13}. Briefly, the main difference between the several QHNN models resides in the activation function. 

In this paper, we consider a quaternion-valued activation function $\sigma$ given by \eqref{eq:sigma} whose output is obtained by normalizing its argument to length one  \cite{valle14bracis,kobayashi16a}. 
The resulting network, referred to as the {\em continuous-valued quaternionic Hopfield neural network} (CV-QHNN), can be implemented and analyzed more easily than multistate quaternionic Hopfield neural network models \cite{isokawa13}. Furthermore, as far as we know, it is the unique version of the Hopfield network on unit quaternions that always yields a convergent sequence in the asynchronous update mode under the usual conditions on the synaptic weights \cite{valle18tnnls}.

The CV-QHNN is defined as follows: Let $w_{ij} \in \mathbb{H}$ denotes the $j$th quaternionic synaptic weight of the $i$th neuron of a network with $n$ neurons. Also, let the state of the CV-QHNN at time $t$ be represented by a column quaternion-valued vector $\vetx(t) = [x_1(t),\ldots,x_n(t)]^T \in \mathbb{S}^n$, that is, the unit quaternion $x_i(t) = x_{i0}(t) + x_{i1}(t)\ii + x_{i2}(t) \jj + x_{i3}(t)\kk$ corresponds to the state of the $i$th neuron at time $t$. Given an initial state (or input vector) $\vetx(0) = [x_1,\ldots,x_n]^T \in \mathbb{S}^n$, the CV-QHNNs defines recursively the sequence of quaternion-valued vectors $\vetx(0),\vetx(1),\vetx(2), \ldots$ by means of the equation
\bb \label{eq:update} x_j(t+1) = \begin{cases} \sigma\left(a_j(t)\right), & 0<|a_j(t)|<+\infty, \\ x_j(t), & \mbox{otherwise}, \end{cases} \ee
where
\bb \label{eq:hopfield} a_i(t)=\sum_{j=1}^n w_{ij} x_j(t), \ee
is the activation potential of the $i$th neuron at iteration $t$. In analogy with the traditional real-valued bipolar Hopfield network, the sequence produced by \eqref{eq:update} and \eqref{eq:hopfield} in an asynchronous update mode is convergent for any initial state $\vetx(0) \in \mathbb{S}^n$ if the synaptic weights satisfy \cite{valle14bracis}:
\bb \label{eq:converg} w_{ij}=\bar{w}_{ji} \qeq w_{ii} \geq 0, \quad \forall i,j \in \{1,\ldots,n\}. \ee 
Here, the inequality $w_{ii} \geq 0$ means that $w_{ii}$ is a non-negative real number.
Moreover, the synaptic weights of a QHNN are usually determined using either the correlation or projection rule \cite{isokawa13}. Both correlation and projection rule yield synaptic weights that satisfy \eqref{eq:converg}.

Consider a fundamental memory set $\mathcal{U}=\{\vetu^1,\ldots,\vetu^p \}$, where each $\vetu^\xi = [u_1^\xi,\ldots,u_n^\xi]^T$ is a quaternion-valued column vector whose components $u_i^\xi=\quat{u^\xi_i}$ are unit quaternions. In the quaternionic version of the {\em correlation rule}, also called {\em Hebbian learning} \cite{isokawa13}, the synaptic weights are given by
\bb \label{eq:correlation} w_{ij}^c = \frac{1}{n} \sum_{\xi=1}^p u_i^\xi \bar{u}_j^\xi, \quad \forall i,j \in \{1,2,\ldots,n\}. \ee  
Unfortunately, such as the real-valued correlation recording recipe, the quaternionic correlation rule is subject to the cross-talk between the original vectors $\vetu^1,\ldots,\vetu^p$. In contrast, the {\em projection rule}, also known as the {\em generalized-inverse recording recipe}, is a non-local storage prescription that can suppress the cross-talk effect between $\vetu^1,\ldots,\vetu^p$ \cite{kanter87}. Formally, in the projection rule the synaptic weights are defined by 
\bb \label{eq:projection} w_{ij}^p = \frac{1}{n} \sum_{\eta=1}^p \sum_{\xi=1}^p u_i^\eta c_{\eta \xi}^{-1} \bar{u}_j^\xi, \ee  
where $c^{-1}_{\eta\xi}$ denotes the $(\eta,\xi)$-entry of the quaternion-valued inverse of the matrix $C \in \mathbb{H}^{p \times p}$ given by
\bb c_{\eta\xi} = \frac{1}{n} \sum_{j=1}^n \bar{u}^\eta_j u_j^\xi =  \frac{1}{n}  \sprod{\vetu^\xi}{\vetu^\eta}, \quad \forall \mu,\nu \in \{1,\ldots,p\}. \ee
It is not hard to show that, if the matrix $C$ is invertible, then $\sum_{j=1}^n w_{ij}^p u_j^\xi = u_i^\xi$ for all $\xi = 1,\ldots,p$ and $i=1,\ldots,n$ \cite{isokawa13}. Therefore, all the fundamental memories are fixed points of the CV-QHNN with the projection rule. On the downside, the projection rule requires the inversion of a $p\times p$ quaternion-valued matrix. 

\section{Quaternion-Valued Recurrent Correlation Neural Networks} \label{sec:QRCNNs}

Recurrent correlation neural networks (RCNNs), formerly known as recurrent correlation associative memories (RCAMs), have been introduced in 1991 by Chiueh and Goodman for the storage and recall of $n$-bit vectors \cite{chiueh91}. The RCNNs have been generalized for the storage and recall of complex-valued and quaternion-valued vectors \cite{valle14nnB,valle18wcci}. In the following, we briefly review the quaternionic recurrent neural networks (QRCNNs). Precisely, to pave the way for the development of the new models introduced in the next section, let us derive the QRCNNs from the correlation-based CV-QHNN described by \eqref{eq:update}, \eqref{eq:hopfield}, and \eqref{eq:correlation}.

Consider a fundamental memory set $\mathcal{U}=\{\vetu^1,\ldots,\vetu^p \} \subset \mathbb{S}^n$. Using the synaptic weights $w_{ij}^c$ given by \eqref{eq:correlation}, we conclude from \eqref{eq:hopfield} that the activation potential of the $i$th neuron at iteration $t$ of the correlation-based CV-QHNN satisfies
\begin{align*}
    a_i(t) &= \sum_{j=1}^n w_{ij}^c x_j(t) = \sum_{j=1}^n \left[\frac{1}{n} \sum_{\xi=1}^p u_i^\xi \bar{u}_j^\xi \right] x_j(t) \\
    &= \sum_{\xi=1}^p u_i^\xi \left[\frac{1}{n} \sum_{j=1}^n \bar{u}_j^\xi x_j(t) \right] \\ &= \sum_{\xi=1}^p u_i^\xi \left[\frac{1}{n} \sprod{\vetx(t)}{\vetu^\xi} \right].
\end{align*} 
In words, the activation potential $a_i(t)$ is given by a weighted sum of $u_i^1,\ldots,u_i^p$. Moreover, the weights are proportional to the inner product between the current state $\vetx(t)$ and the fundamental memory $\vetu^\xi$. 

In the QRCNN, the activation potential $a_i(t)$ is also given by a weighted sum of $u_i^1,\ldots,u_i^p$. The weights, however, are given by function of the real part of the inner product $\sprod{\vetx(t)}{\vetu^\xi}$. Precisely, let $f:[-1,1]\rightarrow \R$ be a (real-valued) continuous and monotone non-decreasing function. Given a quaternionic input vector $\vetx(0) = [x_1(0),\ldots,x_N(0)]^T \in \mathbb{S}^N$, a QRCNN defines recursively a sequence $\{\vetx(t)\}_{t\geq 0}$ of quaternion-valued vectors by means of \eqref{eq:update} where the activation potential of the $i$th output neuron at time $t$ is given by
\bb \label{eq:QRCNN} a_i(t)=\sum_{\xi=1}^p w_\xi(t) u_i^\xi, \quad \forall i=1,\ldots,n,\ee
with
\bb \label{eq:weights} w_\xi(t)= f\left(\frac{1}{n}\re{\sprod{\vetx(t)}{\vetu^\xi}}\right), \quad \forall \xi \in 1,\ldots,p. \ee

Examples of QRCNNs include the following straightforward quaternionic generalizations of the bipolar RCNNs:
\begin{enumerate}
 \item The {\em correlation QRCNN} or {\em identity QRCNN} is obtained by considering in \eqref{eq:weights} the identity function $f_i(x)=x$. 
 \item The {\em high-order QRCNN}, which is determined by the function \bb \label{eq:f_h} f_h(x;q)=(1+x)^q, \quad q>1. \ee
 \item The {\em potential-function QRCNN}, which is obtained by considering in \eqref{eq:weights} the function 
 \bb \label{eq:f_p} f_p(x;L)=\frac{1}{(1-x+\varepsilon_p)^L}, \quad L \geq 1, \ee 
 where $\varepsilon_p>0$ is a small valued introduced to avoid a division by zero when $x=1$. %Like our previous works, the value $\varepsilon_p=\sqrt{\epsilon_{mach}}$, where $\epsilon_{mach}$ denotes the machine floating-point relative accuracy, was used in this paper \cite{valle14nnB,valle18wcci}. 
 \item The {\em exponential QRCNN}, which is determined by an exponential \bb \label{eq:f_e} f_e(x;\alpha)=e^{\alpha x}, \quad \alpha>0. \ee
\end{enumerate}
Note that QRCNNs generalize both bipolar and complex-valued RCNNs \cite{chiueh91,valle14nnB}. Precisely, the bipolar and the complex-valued models are obtained by considering vectors $\vetx=[x_1,\ldots,x_n]^T \in \mathbb{S}^n$ whose components satisfy respectively $x_j = {x_j}_0 + 0\ii + 0\jj + 0\kk$ and $x_j = {x_j}_0+{x_j}_1\ii+0\jj+0\kk$ for all $j=1,\ldots,n$. Furthermore, the correlation QRCNN generalizes the traditional bipolar correlation-based Hopfield neural network but it does not generalize the correlation-based CV-QHNN. Indeed, in contrast to the correlation-based CV-QHNN, the correlation QRCNN uses only the real part of the inner product $\sprod{\vetx(t)}{\vetu^\xi}$. 

Finally, we would like to point out that,  independently of the initial state $\vetx(0) \in \mathbb{S}^n$, a QRCNN model always yields a convergent sequence $\{\vetx(t)\}_{t \geq 0}$. Therefore, QRCNNs are potential models to implement associative memories. Morevoer, due to the non-linearity of the activation functions of the hidden neurons, the high-order, potential-function, and exponential QRCNNs may overcome the rotational invariance problem found on quaternionic Hopfield neural network \cite{kobayashi16a}. On the downside, such as the correlation-based quaternionic Hopfield network, QRCNNs may suffer from cross-talk between the fundamental memories $\vetu^1,\ldots,\vetu^p$. Inspired by the projection rule, the next section introduces improved models which overcome the cross-talk problem of the QRCNNs.

\section{Quaternion-Valued Recurrent Projection Neural Networks}
\label{sec:QRPNNs}

Quaternion-valued recurrent projection neural networks (QRPNNs) combine the main idea behind the projection rule and the QRCNN models to yield high capacity associative memories. Specifically, using the synaptic weights $w_{ij}^p$ given by \eqref{eq:projection}, the activation potential of the $i$th neuron at time $t$ of the projection-based CV-QHNN is
\begin{align*}
    a_i(t) &= \sum_{j=1}^n w_{ij}^p x_j(t) 
    = \sum_{j=1}^n \left[\frac{1}{n} \sum_{\eta=1}^p \sum_{\xi=1}^p u_i^\eta c_{\eta \xi}^{-1} \bar{u}_j^\xi \right] x_j(t) \\
    & = \sum_{\eta=1}^p \sum_{\xi=1}^p u_i^\eta  c_{\eta \xi}^{-1} \left[\frac{1}{n} \sum_{j=1}^n \bar{u}_j^\xi x_j(t) \right] \\
    & = \sum_{\xi=1}^p \left( \sum_{\eta=1}^p u_i^\eta c_{\eta \xi}^{-1} \right) \left[\frac{1}{n} \sprod{\vetx(t)}{\vetu^\xi} \right].
\end{align*} 
In analogy to the QRCNN, we replace the term proportional to the inner product between $\vetx(t)$ and $\vetu^\eta$ by the weight $w_\xi(t)$ given by \eqref{eq:weights}. Accordingly, we define $c^{-1}_{\eta \xi}$ as the $(\eta,\xi)$-entry of the inverse of the real-valued matrix $C \in \mathbb{R}^{p \times p}$ given by
\bb \label{eq:Cmatrix} c_{\eta\xi} = f \left( \frac{1}{n}  \sprod{\vetu^\xi}{\vetu^\eta} \right), \quad \forall \eta,\xi \in \{1,\ldots,p\}.  \ee
Furthermore, to simplify the computation, we define 
\bb \label{eq:vs} v_i^\xi = \sum_{\eta=1}^p u_i^\eta c_{\eta \xi}^{-1},\ee 
for all $i=1,\ldots, n$ and $\eta = 1,\ldots,p$. Thus, the activation potential of a QRPNN is given by
\bb \label{eq:QRPNN} a_i(t)=\sum_{\xi=1}^p w_\xi(t) v_i^\xi, \quad \forall i=1,\ldots,n.\ee

Concluding, given a fundamental memory set $\mathcal{U} = \{\vetu^1,\ldots,\vetu^p\} \subset \mathbb{S}^n$, define the $p \times p$ real-valued matrix $C$ by means of \eqref{eq:Cmatrix} and the compute the quaternion-valued vectors $\vetv^1,\ldots,\vetv^p$ using \eqref{eq:vs}. Like the QRCNN, given an input vector $\vetx(0)\in \mathbb{S}^n$, a QRPNN yields the sequence $\{\vetx(t)\}_{t\geq 0}$ by means of \eqref{eq:update} where the activation potential of the $i$th output neuron at time $t$ is given by \eqref{eq:QRPNN} with $w_\xi(t)$ defined by \eqref{eq:weights}. 

Let us show that QRPNNs overcome the cross-talk problem of the QRCNNs if the matrix $C$ is invertible. Given a fundamental memory set $\mathcal{U}=\{\vetu,\ldots,\vetu^p\}$, let us assume that the matrix $C$ given by \eqref{eq:Cmatrix} is invertible. Suppose that a QRPNN is initialized at a fundamental memory, that is, $\vetx(0)=\vetu^\gamma$ for some $\gamma \in \{1,\ldots,p\}$. From \eqref{eq:weights} and \eqref{eq:Cmatrix}, we conclude that 
\[ \vetw_\xi(0) = f \big( \re{\sprod{\vetu^\gamma}{\vetu^\xi}}/n \big) = c_{\xi \gamma}, \quad \forall \xi =1,\ldots,p.\]
Furthermore, from \eqref{eq:QRPNN} and \eqref{eq:vs}, we obtain the following identities for any $i \in \{1,\ldots,n\}$:
\begin{align*}
    a_i(0) &= \sum_{\xi=1}^p w_\xi(0) v_i^\xi 
    = \sum_{\xi=1}^p \left( \sum_{\eta=1}^p u_i^\eta c_{\eta \xi}^{-1} \right) c_{\xi \gamma} \\
    &= \sum_{\eta=1}^p u_i^\eta \left( \sum_{\xi=1}^p  c_{\eta \xi}^{-1} c_{\xi \gamma}  \right) 
    = \sum_{\eta=1}^p u_i^\eta \delta_{\eta \gamma} = u^\gamma_i,
\end{align*}
where $\delta_{\eta\gamma}$ is the Kronecker delta, that is, $\delta_{\eta \gamma} = 1$ if $\eta = \gamma$ and $\delta_{\eta \gamma}=0$ if $\eta \neq \gamma$. Hence, from \eqref{eq:update}, we conclude that the fundamental memory $\vetu^\gamma$ is a fixed point of the QRPNN if the matrix $C$ is invertible.

% Alternatively, using a matrix-vector notation, a dynamic of a QRPNN is described as follows: Let $U = [\vetu^1,\ldots,\vetu^p] \in \mathbb{S}^{n \times p}$ be the quaternion-valued matrix whose columns corresponds to the fundamental memories. Define the real-valued matrix $C \in \mathbb{R}^{p \times p}$ and the quaternion-valued matrix $V \in \mathbb{H}^{n \times p}$ by means of the equations
% \bb C= f(U^* U/n) \quad \mbox{and} \quad V = U C^{-1},\ee
% where the $f$ is evaluated in an entry-wise manner. Given the initial state $\vetx(0)$, a QRCNN defines recursively
% \bb \vetw(t) = f\big(\re{U^* \vetx(t)}/n\big), \ee and \bb \vetx(t+1) = \sigma\big(V \vetw(t)\big), \ee
% where $f$ and $\sigma:\mathbb{H}^* \to \mathbb{S}$ are evaluated in a component-wise manner. Like the QRCNN, a QRPNN is also implemented by the fully connected two layer neural network with $p$ hidden neurons shown in Fig. \ref{fig:topology}b). The difference between the QRCNN and the QRPNN is the synaptic weight matrix of the output layer. In other words, they differ in the way the real-valued $\vetw(t)$ is decoded to yield the next state $\vetx(t)$. From the computational point of view, although the storing phase of the QRCNN requires $\mathcal{O}(p^3 + n p^2)$ operations to compute $C^{-1}$ and the matrix $V$, this model may settle down at an equilibrium faster than a QRCNN. More importantly, as we shall confirm in the computational experiments in the following section, the QRPNN indeed overcomes the cross-talk between the fundamental memories and exhibit a better noise tolerance than the QRCNNs.

Like the QRCNN, the identity mapping and the functions $f_h$, $f_p$, and $f_e$ given by \eqref{eq:f_h}, \eqref{eq:f_p}, and \eqref{eq:f_e} are used to define respectively the {\em projection or identity QRPNN}, the {\em high-order QRPNN}, the {\em potential-function QRPNN}, and the {\em exponential QRPNN}. Note that the projection QRPNN generalizes the traditional bipolar projection-based Hopfield neural network. The projection QRPNN, however, does not generalize the projection-based CV-QHNN because the former uses only the real part of the inner product between $\vetx(t)$ and $\vetu^\xi$. In fact, in contrast to the projection-based CV-QHNN, the design of a QRPNN does not require the inversion of a quarternion-valued matrix but only the inversion of a real-valued matrix. 

%is defined by \eqref{eq:update} with the activation function \eqref{eq:QRPNN}. Now, using the formulation in \eqref{eq:QRPNN} it is possible to asseverate two important facts, associated to the generalized property when the QRPNN model works in binary spaces as a recurrent projection neural network (RPNN). The first is that the RPNN model corresponds to the HNNP model when $f$ is the identity function. On the other hand, the ECAM model is an instance of the RPNN model. This occurs when in the RPNN model $f$ is an exponential function in \eqref{eq:Cmatrix} and the weighting function belong to the family of \eqref{eq:f_e} for large values of $\alpha$.

\begin{figure*}[t]
    \begin{tabular}{cc} 
    a) Real-valued associative memories & 
    b) Quaternion-valued associative memories \\ 
    \includegraphics[width=1\columnwidth]{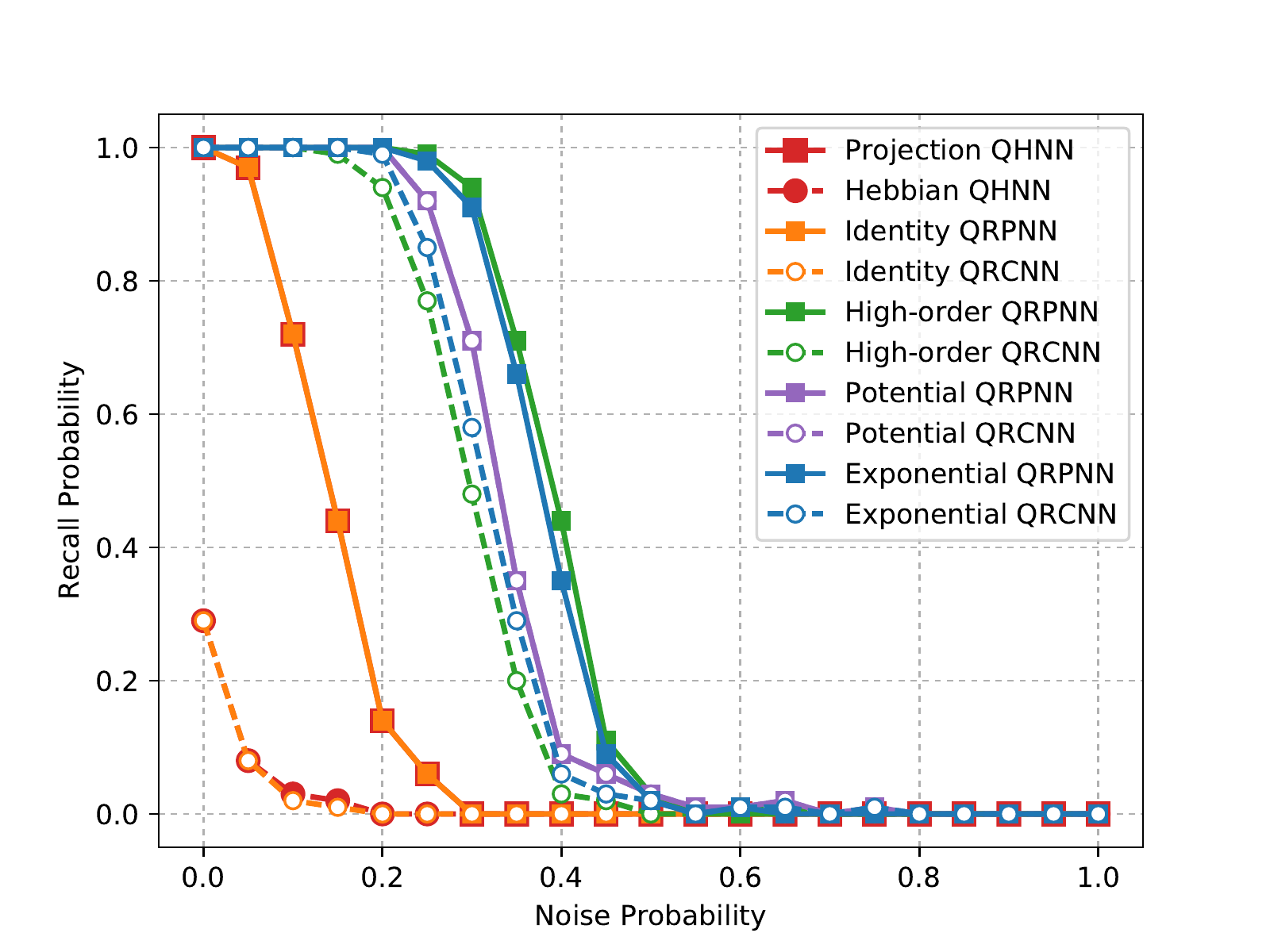} &
    \includegraphics[width=1\columnwidth]{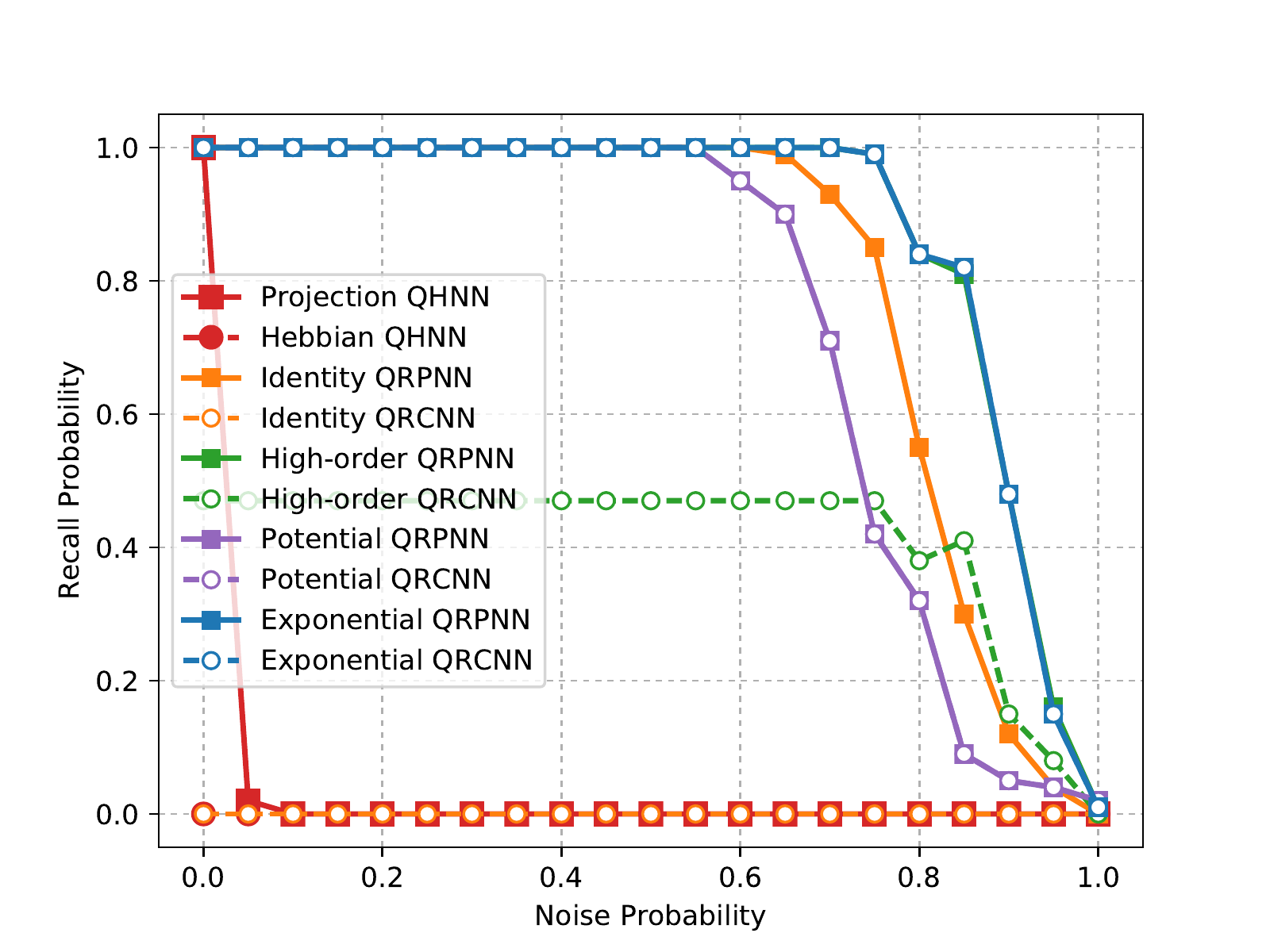}
    \end{tabular}
    \caption{Recall probability by the noise intensity introduced in the input vector.}
    \label{fig:Example1}
\end{figure*}

\begin{example}[Real-valued Associative Memories]  \label{ex:Example1}
Let us compare the storage capacity and noise tolerance of the new RPNNs designed for the storage of $p=36$ randomly generated bipolar (real-valued) vectors of length $n=100$ with the original RCNNs and Hopfield neural netwokrs (HNNs). Precisely, we consider projection and Hebbian Hopfield neural network, the identity, high-order, potential-function, and exponential RCNNs and RPNNs with parameters $q=5$, $L=3$, and $\alpha=4$. 
To this end, the following steps have been performed 100 times: 
\begin{enumerate}
\item We synthesized associative memories designed for the storage and recall of a randomly generated fundamental memory set $\mathcal{U}=\{\vetu^1,\ldots,\vetu^p\}$, where $\mbox{Pr}[u_i^\xi = 1] = \mbox{Pr}[u_i^\xi = -1]  = 0.5$  for all $i=1,\ldots,n$ and $\xi=1,\ldots,p$. 
\item We probed the associative memories with an input vector $\vetx(0)=[x_1(0),\ldots,x_n(0)]^T$ obtained by reversing some components of $\vetu^1$ with probability $\pi$, i.e., $\mbox{Pr}[x_i(0) = -u_i^1] = \pi$ and $\mbox{Pr}[x_i(0) = u_i^1] = 1-\pi$, for all $i$ and $\xi$. 
\item The associative memories have been iterated until they reached a stationary state or completed a maximum of 1000 iterations. A memory model succeeded to recall a stored item if the output equals $\vetu^1$. 
\end{enumerate} 
Fig. \ref{fig:Example1}a) shows the probability of an associative memory to recall a fundamental memory by the probability of noise introduced in the initial state. Note that the projection HNN coincides with the identity RPNN. Similarly, the Hebbian HNN coincides with the identity RCNN. Note that the RPNNs always succeeded to recall undistorted fundamental memories (zero noise probability). The high-order, potential-function, and exponential RCNNs also succeeded to recall undistorted fundamental memories. Nevertheless, the recall probability of the high-order and exponential RPNNs are greater than or equal to the recall probability of the corresponding RCNNs. In other words, the RPNNs exhibit better noise tolerance than the corresponding RCNNs. The potential-function RCNN and RPNN yielded similar recall probabilities.
\end{example}

\begin{example}[Quaternion-valued Associative Memories] \label{ex:Example2}
Let us now investigate the storage capacity and noise tolerance of the new asssociative memory models for the storage and recall of $p=36$ randomly generated quaternion-valued vectors of length $n=100$. In this example, we considered projection and Hebbian quaternion-valued Hopfield neural network (QHNN), the identity, high-order, potential-function, and exponential QRCNNs and QRPNNs with parameters $q=20$, $L=3$, and $\alpha=14$. These parameters have been determined so that the QRCNNs have more than 50\% probability to recall undistorted fundamental memories.
In analogy to the previous example, the following steps have been performed 100 times: 
\begin{enumerate}
\item We synthesized associative memories designed for the storage and recall of uniformly distributed fundamental memories $\mathcal{U}=\{\vetu^1,\ldots,\vetu^p\}$. Formally, we defined $u_i^\xi = \mathtt{RandQ}$ for all $i=1,\ldots,n$ and $\xi=1,\ldots,p$ where
\[ \mathtt{RandQ} = (\cos \phi+\ii\sin\phi)(\cos\psi+ \kk\sin \psi)(\cos \theta+\jj \sin \theta), \]
is a randomly generated unit quaternion obtained by sampling angles $\phi \in [-\pi,\pi)$, $\psi \in [-\pi/4,\pi/4]$, and $\theta \in [-\pi/2,\pi/2]$ using an uniform distribution. 
\item We probed the associative memories with an input vector $\vetx(0)=[x_1(0),\ldots,x_n(0)]^T$ obtained by replacing some components of $\vetu^1$ with probability $\pi$ by an uniformly distributed component, i.e., $\mbox{Pr}[x_i(0) = \mathtt{RandQ}] = \pi$ and $\mbox{Pr}[x_i(0) = u_i^1] = 1-\pi$, for all $i$ and $\xi$. 
\item The associative memories have been iterated until they reached a stationary state or completed a maximum of 1000 iterations. The memory model succeeded if the output equals the fundamental memory $\vetu^1$. 
\end{enumerate} 
Fig. \ref{fig:Example1}b) shows the probability of a quaternion-valued associative memory to recall a fundamental memory by the probability of noise introduced in the initial state. 
As expected, the QRPNNs always succeeded to recall undistorted fundamental memories. The potential-function and exponential QRCNNs also succeeded to recall undistorted fundamental memories. Indeed, the potential-function QRCNN and QRPNNs yielded the same recall probability. The noise tolerance of the exponential QRCNN and QRPNN also coincided. Nevertheless, the recall probability of the high-order QRPNN is greater than the recall probability of the corresponding QRCNNs. Furthermore, in contrast to the real-valued case, the projection QHNN differs from the identity QRPNN. In fact, the noise tolerance of the identity QRPNN  is far greater than the noise tolerance of the projection QHNN.
% \begin{figure}[t]
%     \centering
%     \includegraphics[width=1\columnwidth]{Figures/Example2}
%     \caption{Recall probability of quaternion-valued associative memories by the noise intensity introduced in the input vector.}
%     \label{fig:Example2}
% \end{figure}
\end{example}

\section{Concluding Remarks} \label{sec:concluding}

In this paper, we introduced the quaternion-valued recurrent projection associative memories (QRPNNs) by combining the projection rule with the quaternion-valued recurrent correlation associative memories (QRCNNs). In contrast to the QRCNNs, QRPNNs always exhibit optimal storage capacity. In fact, preliminary computational experiments (Examples \ref{ex:Example1} and \ref{ex:Example2}) show that the storage capacity and noise tolerance of QRPNNs (including real-valued case) are greater than or equal to the storage capacity and noise tolerance of their corresponding QRCNNs. In the future, we plan to investigate further the noise tolerance of the QRPNNs. We also intent to address the performance of the new associative memories for pattern reconstruction and classification.

% The main change in the QRPNN model respect to the other models, is the construction of the storage matrix. The derived $\mathcal{C}$ matrix is the main difference which take the essence of the projection rule and the exponentiation, combining both concepts to generalize. The QRPNN model keeps an stable behavior for small and large values of $\alpha$, making this model more accessible to real applications, because it avoid the overflow produced by taking large values for this parameter. This comparison is presented in Fig. \ref{f4} where the QRPNN model keep stable in relation to the Potential-function QRCNN which can not take large values of the exponential parameter $\alpha$. Moreover, we generalized the Hopfield neural network HNN
% with the projection rule \cite{personnaz85} as well as the exponential correlation neural network of Chiueh and Goodman \cite{chiueh91} for particular choices of the weighting function. The QRPNN model overcome the cross-talk problem of the QRCNNs if the matrix $C$ is invertible, because if this condition is satisfied then every fundamental memory $\vetu^\gamma$ is a fixed point of the QRPNN. In relation to the model dynamic behavior in the binary case, for small values of $\alpha$ we could verify the main difference between the ECAM and RPNN, showed in the noise response in Fig. \ref{f3}. In this sense, the RPNN model has more stability for these special cases. 

% \bibliographystyle{IEEEtran}
% \bibliography{referencesBRACIS19}


\end{document}